# Enrichment of OntoSenseNet: Adding a Sense-annotated Telugu lexicon

Sreekavitha Parupalli and Navjyoti Singh

Center for Exact Humanities (CEH)
International Institute of Information Technology, Hyderabad, India
sreekavitha.parupalli@research.iiit.ac.in
navjyoti@iiit.ac.in

**Abstract** The paper describes the enrichment of OntoSenseNet- a verb-centric lexical resource for Indian Languages. This resource contains a newly developed Telugu-Telugu dictionary. It is important because native speakers can better annotate the senses when both the word and its meaning are in Telugu. Hence efforts are made to develop a soft copy of Telugu dictionary. Our resource also has manually annotated gold standard corpus consisting 8483 verbs, 253 adverbs and 1673 adjectives. Annotations are done by native speakers according to defined annotation guidelines. In this paper, we provide an overview of the annotation procedure and present the validation of our resource through inter-annotator agreement. Concepts of sense-class and sense-type are discussed. Additionally, we discuss the potential of lexical sense-annotated corpora in improving word sense disambiguation (WSD) tasks. Telugu WordNet is crowd-sourced for annotation of individual words in synsets and is compared with the developed sense-annotated lexicon (OntoSenseNet) to examine the improvement. Also, we present a special categorization (spatio-temporal classification) of adjectives.

## 1 Introduction

Lexically rich resources form the foundation of all natural language processing (NLP) tasks. Maintaining the quality of resources is thus a high priority issue [5]. Hence, it is important to enhance and maintain the lexical resources of any language. This is of significantly more importance in case of resource poor languages like Telugu [19].

WordNet is a vast repository of lexical data and it is widely used for automated sense-disambiguation, term expansion in IR systems, and the construction of structured representations of document content [12]. First WordNet among the Indian languages was developed for Hindi. WordNets for 16 other Indian languages are built from Hindi WordNet applying expansion approach [1].

WSD can be characterized as a task that emphasizes on evaluating the right sense of a word in its particular context. It is a critical pre-processing step in data extraction, machine translation, question answering systems and numerous other NLP tasks. Vagueness in word sense emerges when a specific word has multiple conceivable senses. Finding the right sense requires exhaustive information of words. This additional information be call as *intentional* meaning of the word. Meaning can be discussed as sense (intensional meaning) and reference (extensional meaning)[6]. The meaning of a word, from ontological viewpoint, can be understood based on its participation in classes, events and relations. We use a formal ontology that is developed to computationally manipulate language at the level of meanings which have an intrinsic form [13].

The paper is organized as follows. Section 2 discusses available lexical resources for Telugu and several types of WSD tasks that were previously developed. Section 3 describes our dataset and shows the statistics of available lexical resources.Section 4 talks about the ontological classification that was formalized for the annotation purpose by [13]. Section 5 describes annotation guidelines and explains the procedure of manual annotation by expert native speakers. Section 9 concludes the paper and section 10 presents the scope of future work in the domain.

IAST based transliteration[1] for Telugu script has been employed in the paper.

## 2 Related Work

Before understanding the tasks that are performed, it is important to understand and analyze the available resource thoroughly. Hence this section discusses the previous work that was done in this domain.

### 2.1 Telugu WordNet

Telugu WordNet is developed as a part of IndoWordNet[2] at CFILT [2], which is considered as the most exhaustive set of multilingual lexical assets for Indian languages. It consists of 21091 synsets in total. This total includes 2795 verb synsets, 442 adverb synsets, 5776 adjective synsets. Telugu WordNet captures several other semantic relations such as hypernymy, hyponymy, holonymy, meronymy, antonymy. For every word in the dictionary it provides synset ID, parts-of-speech (POS) tag, synonyms, gloss, example statement, gloss in Hindi, gloss in English. An example of an entry in the IndoWordNet database is shown in Figure 1.

### 2.2 Ontological issues in WordNet

WordNet is a language specific resource and it varies from language to language. However, any WordNet can be considered an ontology through the hypernymy-hyponymy relations that are present in it. WordNet of any language leaves a few loop holes that other ontologies can fill [1]. By summarizing the following [11], [7], [15], we state the four major problems:

---

[1] http://www.learnsanskrit.org/tools/sanscript
[2] http://www.cfilt.iitb.ac.in/indowordnet/index.jsp

| | |
|---|---|
| Number of Synset for "అడవి" : 6 | showing 1 / 6 |
| Synset ID : 2551 | POS : noun |
| Synonyms : | అడవి, అటవి, అరణ్యం, కాన, |
| Gloss : | దట్టమైన చెట్లపొదలతో క్రూరమృగాలతో ఉండే స్థలం |
| Example statement : | "పురాతన కాలంలో ఋుషులు-మునులు అడవిలో నివాసం ఉండేవారు." |
| Gloss in Hindi : | यह स्थान जहाँ बहुत दूर तक पेड़-पौधे, झाड़ियाँ आदि अपने आप उगी हों |
| Gloss in English : | land that is covered with trees and shrubs |

Figure 1: Example entry in the IndoWordNet database

- *Confusing concepts with individuals:* WordNet synsets do not distinguish between universal and the particular instance of a concept. For example, both 'aḍavi(forest)' and 'ceṭṭu(tree)' are considered as concept.
- *Lexical gap:* A language may not have an indigenous lexeme to describe a concept. For example, vehicles can be divided into two classes, 1) Vehicles that run on the road and 2) Vehicles that run on the rail, but language may not have specific words to describe these classes.
- *Confusion between object level and meta level concept:* The synset abstraction seems to include both object-level concepts, such as Set, Time, and Space, and meta-level concepts, such as Attribute and Relation [7].
- *Heterogeneous levels of generality:* Two hyponyms of a concept may represent different level of generality. For example, as a hyponymy of concept 'aḍavi (forest)', there is a general concept 'podalu (bushes)' and a more specific concept 'kalabanda (aloe vera)', a medicinal plant. We are induced to consider the formers as types and the latter as roles. In other words, we discover that, if at first sight some synsets sound intuitively too specific when compared to their siblings, from a formal point of view, we may often explain their "different generality" by means of the distinction between types and roles [8].

## 2.3 Variants of WSD

WSD is broadly categorized into two types [3]:

- *Target Word WSD:* The target WSD system disambiguates a restricted set of target words, usually one per sentence. Supervised approaches are generally used for WSD where a tagged corpus is used to train the model. This trained model is then used to disambiguate the words in the target document.
- *All Word WSD:* The all word WSD system disambiguates all open-class words in the target document. Supervised approaches face the problem of data sparseness and it is not always possible to have a large tagged corpus for training. Hence, unsupervised methods are preferred in the case of all word WSD.

### 2.4 Approaches for Word Sense Disambiguation

WSD approaches are often classified according to the main source of knowledge used in sense differentiation. We are listing a few as discussed in [3].

- Supervised WSD Approaches: Supervised methods formulate WSD as a classification problem. The senses of a word represent classes and a classifier assigns a class to each new instance of a word. Any classifier from the machine learning literature can be applied. In addition to a dictionary, these algorithms need at least one annotated corpus, where each appearance of a word is tagged with the correct sense.
- Unsupervised WSD Approaches: Creating annotated corpus for all language-domain pairs is impracticable looking at the amount of time and money required. Unsupervised methods have the potential to overcome the new knowledge acquisition bottlenecK. These methods are able to induce word senses from training text by clustering word occurrences and then classifying new occurrences into the induced clusters/senses.
- Knowledge Based WSD Approaches: WSD heavily depends on knowledge and this knowledge must be in the machine readable format. There are various structures designed for this purpose like tagged and untagged corpora, machine-readable dictionaries, ontologies, etc. The main use of lexical resources in WSD is to associate senses with words. Here, selectional restrictions, overlap of definition text, and semantic similarity measures are used for knowledge based WSD.

## 3 Data Collection

Telugu is the second most spoken language in India. It is one of the twenty-two official languages of the Republic of India and the official language of the states of Telangana and Andhra Pradesh. Telugu has a vast and rich literature dating back to many centuries [9].

However, there is no generally accessible dictionary reference till date. In this work, a Telugu lexicon was created manually from 'శ్రీసూర్యరాయాంధ్రతెలుగు నిఘంటువు (Srī sūryarāyāṃdhra Telugu nighaṃṭuvu)' which has 8 volumes in total [14]. Nearly 21,000 root words alongside their their meanings were recorded. The resource is developed to enrich OntoSenseNet[3] with addition of regional language resources. For each word extracted, based on its meaning, sense was identified by native speakers of language. We are presenting some statistics of available resources in Table 1. There are around 36,000 words in the dictionary we developed whereas IndoWordNet lists 21,091 words. Even without further analysis and classification we can see that this resource enriches WordNet by adding almost 15,000 words. This was the motivation to start this work. Nouns are still being added to our resource. Telugu-Hindi and English-Telugu dictionaries are available[4].

---

[3] http://ceh.iiit.ac.in/lexical_resource/index.html
[4] https://ltrc.iiit.ac.in/onlineServices/Dictionaries/Dict_Frame.html

| Resource | Verbs | Adverbs | Adjectives |
|---:|---:|---:|---:|
| OntoSenseNet | 8483 | 253 | 1673(In progress) |
| Telugu WordNet [5] | 2803 | 477 | 5827 |
| Synsets in WordNet | 2795 | 442 | 5776 |
| Telugu-Hindi Dictionary[6] | 9939 | 142 | 1253 |
| English-Telugu Dictionary[7] | 4657 | 1893 | 6695 |

Table 1: Statistics of available lexical resources for Telugu

### 3.1 Validation of the Resource

Cohen's Kappa [4] was used to measure inter-annotator agreement which proves the reliability. The annotations are done by one human expert and it is cross-checked by another annotator who is equally proficient. Both the annotators are native speakers of the language. Verbs and adverbs are randomly selected from our resource for the evaluation sample. The inter-annotator agreement for 500 Telugu verbs is 0.86 and for 100 Telugu adverbs it is 0.94. Validation of the language resource shows high agreement [10]. However further validation of the resource is in progress.

## 4  Ontological Classification Used for Annotation

The formal ontology we used in this paper is proposed by [17]. This is based on various theories given in Indian grammatical tradition. The two main propositions given in Indian grammatical tradition are : (a) All words (noun and verb) in a language can be derived from verbal root (Sanskrit, dhātu). (b) Verbs have operation/process as its predominant element [17]. Theory used in this paper believes that meanings have primitive ontological forms and aims at extensive coverage of language.

### 4.1  Verb

Verbs are considered as the most important lexical and syntactic category of language. Verbs provide relational and semantic framework for its sentences. In a single verb many verbal sense-types can be present and different verbs may share same verbal sense-types. There are seven sense-types of verbs have been derived by collecting the fundamental verbs used to define other verbs [13]. These sense-types are inspired from different schools of Indian philosophies. The seven sense-types of verbs are listed below [16] with their primitive sense along with Telugu examples.

- Means|End - A process which cannot be accomplished without a doer (To do). Examples: *parugettu (run), moyu (carry)*
- Before|After - Every process has a movement in it. The movement maybe a change of state or location (To move). Examples: *pravāhaṁ (flow), oragupovu (lean)*

- Know|Known - Conceptualize, construct or transfer information between or within an animal (To know). Examples: *daryāptu (investigate), vivaraña (explain)*
- Locus|Located - Continuously having (to be in a state) or possessing a quality (To be). Examples: *Ādhārapaḍi (depend), kaṅgāru (confuse)*
- Part|Whole - Separation of a part from whole or joining of parts into a whole. Processes which causes a pain. Processes which disrupt the normal state (To cut). Examples: *perugu (grow), abhivṛddhi (develop)*
- Wrap|Wrapped - Processes which pertain to a certain specific object or category. It is like a bounding (To cover). Examples: *dhariñcaḍaṁ (wear), Āśrayaṁ(shelter)*
- Grip|Grasp - Possessing, obtaining or transferring a quality or object (To have). Examples: *lāgu (grab), vārasatvaṅga (inherit)*

### 4.2 Adverb

Meaning of verbs can further be understood by adverbs, as they modify verbs. The sense-classes of adverbs are inspired from adverb classification in Sanskrit as reported by [13]. Sense-classes with explanation are illustrated with Telugu examples in table 2.

| Sense-Class | Explanation | Example |
|---|---|---|
| Temporal | Adverbs that attributes to sense of time. | akāraṇamu |
| Spatial | Adverbs that attributes to physical space | diguvagā |
| Force | Adverbs that attributes to cause of happening | nikkamu |
| Measure | Adverbs dealing with comparison | niṁḍu |

Table 2: Sense-Class categorization of Adverbs

### 4.3 Adjectives

Like verbs, adjectives are also collocative in nature. [13] identifies 12 sense-types. However these can be reduced to 6 pairs. Sense-Types of adjectives with explanation are illustrated with Telugu examples in table 3.

**4.3.1 Spatio-Temporal Classification of Adjectives** Additional information that we can use for classification are the locational and temporal attributes of adjectives. This can help the machine understand the sense in which a particular adjective is used. In this paper we are proposing three classes, namely, (a) Adjectives dealing with disposition (b) Adjectives of experience (c) Adjectives that talk about the behavior.

| Sense-Type | Explanation | Example |
|---|---|---|
| Locational | Adjectives that universalize or localize a noun | nirdista (specific) |
| Quantity | Adjectives that either qualify cardinal measure or quantify in ordinal-type | okkati (One) |
| Relational | Adjectives that qualify nouns in terms of dependence or dispersal | vistrta (broad) |
| Stress | Adjectives that intensify or emphasis a noun | gatti (strong) |
| Judgement | Adjectives that qualify evaluation or qualify valuation feature of a noun | mamci (good) |
| Property | Adjectives that attribute a nature or qualitative domain of a noun. | nallani (black) |

Table 3: Sense-Type Classification of Adjectives

– *Disposition:* These are the tendencies we have and our habits. We do these unconsciously with not much thought. In ontological terms, this is the trans-temporal categorization.
  Example: mañci vyakti (A good person) Here mañci(good) is used to determine the quality of a man. Here his goodness is reflected in all of his doings. This is an opinion that could be formed after observing him over time and not by judging any one action. Hence it is trans-temporal.
– *Experience:* These could be defined as the adjectives which express the emotion or cognition at any particular moment in time.
  Example: kōpantō unna vyakti (an angry man). This shows the state of the man at that particular point in time.
– *Behavior:* This category of adjectives describe the physical attributes and bodily actions. Hence this is the spatial categorization.
  Example: biggaragā aravaḍaṁ (Loud scream). This describes the action of a person.

This classification is attempted for 400 adjectives in Telugu.

## 5 Annotation Procedure

Every verb, in OntoSenseNet, can have all the seven meaning primitives (sense-types) in it, in various degrees. The degree depends on the usage or popularity of a meaning in a language that leads to a particular sense-type annotation. In our resource we have identified two sense-types for each verb, i.e. primary and secondary. Entire lexicon of verbs and adverbs is classified. However, work is in progress for adjectives. Till date, 1673 adjectives are annotated. All of the annotations are done manually by native speakers of language in accordance with the classification presented in section 4

### 5.1 Enrichment of the Resource

We did not overlook the possibility of existence of unseen words in Telugu WordNet but not in our resource. Out of 2795 verb synsets, we extracted a bag of words which are not present in the resource in developed. We annotated each lexeme of these synsets as an separate entry. We followed similar annotation guidelines for the synsets of WordNet as well. Annotations for this set of lexemes are crowd-sourced and annotations are done following the annotation guidelines by six language experts. We can observe that all the lexemes in synsets (in Telugu wordNet) don't share the same primary sense-type. Another hypothesis is that having sets that share the same primary and secondary sense-types would result in better WSD for tasks like machine translation. However, this hypothesis needs further experimental validation.

### 5.1.1   Adding synsets from WordNet to our resource ID :: 3434
CAT :: verb
CONCEPT :: ప్రతిరోజు సూర్యుడు తూర్పున రావడం (pratiroju sūryuḍu tūrpuna rāvaḍaṃ)
EXAMPLE :: సూర్యుడు తూర్పున ఉదయిస్తాడు (sūryuḍu tūrpuna udayistāḍu)
SYNSET-TELUGU :: ఉదయించు (udayiṃcu), పుట్టు(puṭṭu), పొడతేంచు (poḍateṃcu), అవతరించు (avatariṃcu), ఆవిర్భవించు (āvirbhaviṃcu), ఉద్భవించు (udbhaviṃcu), జనించు (janiṃcu), జనియించు(janiyiṃcu), ప్రభవించు(prabhaviṃcu), వచ్చు (vaccu), ఏతెంచు (eteṃcu)

In synset ID 3434, the verb puṭṭu (birth) is used in the sense of Sun rising in the east. In a sense that sun is taking birth i.e. it conveys that sun came into existence. The primary sense of this would be 'Before|After' as it deals with transition. Secondary sense would be 'Locus|Located' as it shows the state of a sun in the dawn.

However, another (synonymous) word, janiṃcu (birth), in the synset is used to describe the birth of a child. In a sense that a mother gave birth to her child. This process of child-birth needs an agent hence the primary sense becomes 'Means|End' as the action needs agent for its accomplishment. The secondary sense would be 'Part|Whole' as the child was separated from a whole i.e. his mother.

In this example, words from same synset have different primary sense-type. There is a high potential for such occurrences hence each word in the synset was considered as a new entry for the annotation task rather than assigning same primary and secondary sense-type to all the words in a synset.

## 6   Challenges during annotation

A lot of times during annotation, words with confusing sense-types have occurred. For example, consider the words : కోపించు (kopiṃcu); ఎదురుకొలుపు (edurukolupu)

Some words that are in the developed Telugu dictionary are not in use anymore. Many such forgotten words are encountered and even the gloss of the words weren't helpful for the annotation. Such words are left out. For example: అగడాడ (agaḍāḍa): ప్రేలు (prelu), వదరు (vadaru); దరికొను (darikonu) : దహించు (dahiṃcu) , కాల్చు(kālcu)

In some cases, the word is unknown however knowing the gloss helped in the classification task. For example: అక్కటికించు (akkaṭikiṃcu) : కరుణించు(karuṇiṃcu), జాలిపడు (jālipaḍu); అండగొట్టు (aṃḍagoṭṭu) : ఆలస్యము చేయు (ālasyamu ceyu); త్రస్త-రించు (trastariṃcu) : క్రిందపరుచు (kriṃduparucu), అధఃకరించు(adhaḥkariṃcu)

The annotations done involve a lot of time and manual labor hence they are very cost intensive.

## 7 Comparative Analysis

Similar resources have been developed for English and Hindi as well. The differences in the sense distribution of these languages could be due the syntactic and semantic properties of the language. Telugu has many inflections and it is highly agglutinative.

Figure 2 shows the sense-type distribution for English, Hindi and Telugu verbs in OntoSenseNet.

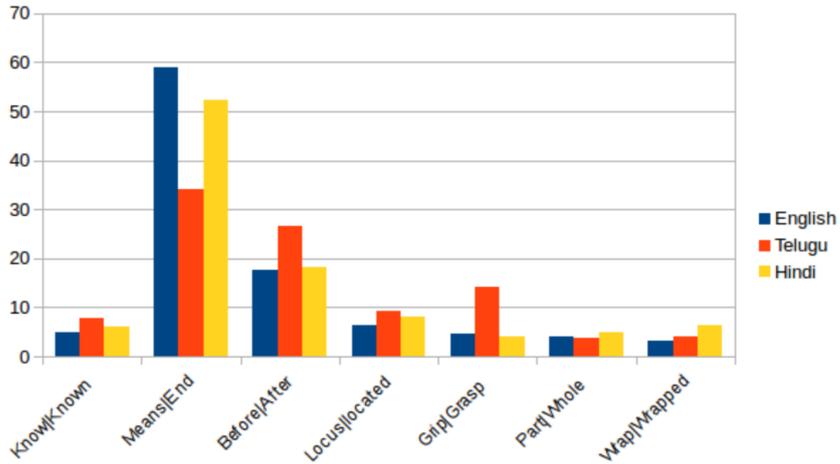

Figure 2: Verb sense-type distribution across langauges

Table 4 shows sense-class distribution of adverbs for OntoSenseNet-English, OntoSenseNet-Hindi and OntoSenseNet-Telugu.

## 8 Adverbial Class Distribution of Verbs

We have extracted all the <Verb, Adverb> and <Adverb, Verb> pairs from the Telugu Wikipedia. In order to acquire these patterns we performed the task of POS tagging[8] on Wikipedia corpus. From the extracted pairs, we noticed that there are comparatively more <Adverb, Verb> pairs than <Verb, Adverb> pairs which align with the structure of Telugu language[18]. 400 verbs and 445 adverbs are annotated according to the formal ontology that is discussed in section 4, A Formal Ontology-based Classification of Lexemes. These words formed about 2000 <Verb, Adverb> and <Adverb, Verb> pairs. Our aim is to study the adverbial class distribution of verbs in Telugu. [13] proves that such annotations help in disambiguating the word senses thus result in improved word-sense disambiguation (WSD) task(s). This is one of the major applications of OntoSenseNet.

| Sense-Class | English | Hindi | Telugu |
|---|---|---|---|
| Temporal | 5.5% | 24.3% | 28.7% |
| Spatial | 2.7% | 13.5% | 12.8% |
| Measure | 39.4% | 32.2% | 31.6% |
| Force | 52.2% | 30% | 26.7% |

Table 4: Adverb Sense-Class Distribution

In table 5, we show the adverbial class distribution of verbs in <Verb, Adverb> and <Adverb, Verb> pairs. Adverbial sense-classes are labeled as columns and sense-types of verbs are labeled as rows. Any cell in the table represents the percentage of a 'sense-class' of adverbs that modify a particular 'sense-type' of verbs.

Column-1 of 5 means 20.0% of 'spatial'; 13.6% of 'temporal'; 18.8% of 'force' and 24.4% of 'measure' sense-classed of adverbs modify 'to know' sense-type of verbs. This shows that majority of the 'to know' verbs are primarily modified by adverbs with 'measure' sense-class. For example : *cālā anipiṃciṃdi* (feel immensely). 'To move', 'to do' verbs are primarily modified by 'spatial', 'force' sense-class of adverbs respectively. Examples are *nerugā māṭlāḍutāḍu*(talk in a straight forward manner), *emoṣanalgā ālocistāḍu* (think emotionally). 'Temporal' sense-class of adverbs can modify all the sense-types of verbs. 'To be' sense-type of verbs is also significantly modified by 'force' sense-class of adverbs. However, 'temporal' and 'measure' sense-classes also seem to show comparable performance in modifying 'to be' sense-type. We can find many such examples in Telugu language.

## 9   Conclusion

In this paper, a manually annotated sense lexicon was developed. Classification was done by expert native Telugu speakers. This sense-annotated resource is an attempt to make machine as intelligible as a human while performing WSD tasks. Hence, without limiting to the word and its meaning, we attempted to convey the sense in which humans understand a sentence. The validation of this resource was done using Cohen's Kappa that showed higher agreement. Further validation and enrichment of the resource is in progress. Classification of WordNet was attempted to see if the proposed classification could improve WSD tasks.

---

[8] https://bitbucket.org/sivareddyg/telugu-part-of-speech-tagger

|          | To Know | To Move | To Do | To Have | To Be  | To Cut | To Bound |
|----------|---------|---------|-------|---------|--------|--------|----------|
| Spatial  | 20.0 %  | 28.5%   | 20 %  | 9.5 %   | 9.5 %  | 8.5%   | 4.0 %    |
| Temporal | 13.6 %  | 22.0 %  | 14.6% | 20.5%   | 20.5 % | 4.4%   | 4.4 %    |
| Force    | 18.8 %  | 21.5 %  | 22.2% | 7.2%    | 22.9%  | 6.0%   | 4.1%     |
| Measure  | 24.4%   | 16.5 %  | 19.5% | 5.2%    | 20.3 % | 7.5%   | 3.8%     |

Table 5: Adverb Sense-Class Distribution in <Verb,Adverb> pairs

## 10   Future Work

Annotations of all available synsets of the WordNet needs to be done to spot the anomalies. The anomalies should be studied to further enrich Telugu WordNet. Tagging of adjectives in still in progress. Supervised WSD approaches are to be implemented by using the OntoSenseNet, sense-annotated corpora. Knowledge based WSD approaches can also benefit largely from such sense-annotated corpora hence such classifiers could be implemented as well. We need to measure the significance of this lexicon in NLP tasks. There are several ontological problems with the WordNet and we are attempting to solve a part of those with the proposed formal ontology. But a far more important question is "How can we know when an ontology is complete?". We hope to arrive at an answer for this question in near future.

## 11   Acknowledgements

This work is part of the ongoing MS thesis in Exact Humanities under the guidance of Prof. Navjyoti Singh. I am immensely grateful to Vijaya Lakshmi for helping me with data collection and annotation process of the whole resource. I would also like to show my gratitude to K Jithendra Babu, Historian & Chairman at Deccan Archaeological and Cultural Research Institute, Hyderabad for providing the hard copy of the dictionary (all the 8 volumes most of which are unavailable currently) and for sharing his pearls of wisdom with us during the course of this research. I thank my fellow researchers from LTRC lab, IIIT-H who provided insight and expertise that greatly assisted the research.

12      Sreekavitha Parupalli and Navjyoti Singh